\documentclass{article}

\usepackage[final]{timeseries_workshop}

\usepackage[utf8]{inputenc} 
\usepackage[T1]{fontenc}    
\usepackage{hyperref}       
\usepackage{url}            
\usepackage{booktabs}       
\usepackage{amsfonts}       
\usepackage{nicefrac}       
\usepackage{microtype}      
\usepackage{xcolor}         
\usepackage{graphicx}
\usepackage{todonotes}

\usepackage{framed}
\usepackage{changepage}  

\definecolor{formalshade}{rgb}{0.95,0.95,1}
\definecolor{darkblue}{rgb}{0.63,0.73,0.86}

\newenvironment{formal}{%
  \MakeFramed{\advance\hsize-\width\FrameRestore}%
  \noindent\hspace{-4.55pt}
  \begin{adjustwidth}{}{3pt}%
  \small
  \vspace{2pt}\vspace{2pt}%
}
{%
  \vspace{2pt}\end{adjustwidth}\endMakeFramed%
}
\usepackage{amsmath}
\usepackage{pifont}

\usepackage{natbib}
\setcitestyle{numbers,open={[},close={]}} 

\title{LLMForecaster: Improving Seasonal Event Forecasts with Unstructured Textual Data}

\author{
  Hanyu Zhang\thanks{Work done while at Amazon} \\
  Georgia Institute of Technology \\
  \texttt{hzhang747@gatech.edu} \\ 
  \And 
  Chuck Arvin\thanks{Corresponding author} \\
  Amazon \\
  \texttt{chuarvin@amazon.com} \\
  \And
  Dmitry Efimov \\
  Amazon \\
  \texttt{defimov@amazon.com}
  \And
  Michael W. Mahoney \\
  Amazon \\
  \texttt{zmahmich@amazon.com} \\
  \And
  Dominique Perrault-Joncas \\
  Amazon \\
  \texttt{joncas@amazon.com} \\
  \And 
  Shankar Ramasubramanian\\
  Amazon \\
  \texttt{sramasub@amazon.com} \\
  \And 
  Andrew Gordon Wilson \\
  Amazon \& NYU \\
  \texttt{wilsmman@amazon.com} \\ 
  \And 
  Malcolm Wolff \\
  Amazon \\
  \texttt{wolfmalc@amazon.com} 
}

\begin{document}

\maketitle

\begin{abstract}
Modern time-series forecasting models often fail to make full use of rich unstructured information about the time series themselves. This lack of proper conditioning can lead to "obvious" model failures; for example, models may be unaware of the details of a particular product, and hence fail to anticipate seasonal surges in customer demand in the lead up to major exogenous events like holidays for clearly relevant products. 
To address this shortcoming, this paper introduces a novel forecast post-processor --- which we call LLMForecaster --- that fine-tunes large language models (LLMs) to incorporate unstructured semantic and contextual information and historical data to improve the forecasts from an existing demand forecasting pipeline. 
In an industry-scale retail application, we demonstrate that our technique yields statistically significantly forecast improvements across several sets of products subject to holiday-driven demand surges.
\end{abstract}

\section{Introduction}
\vspace{-3mm}
Time series forecasting has a broad variety of uses across industry today, including in transportation, weather and retail settings. In modern retail settings, accurate demand forecasts are key to running an efficient supply chain. Improvements in forecast quality directly affect inventory efficiency, reducing stockouts, and enhancing customer satisfaction. 

In recent years, deep neural networks have become a powerful tool for forecasting at scale.
Deep learning models such as Recurrent Neural Networks (RNNs) \cite{hochreiter1997long,fischer2018deep}, Convolutional Neural Networks (CNNs) \cite{lecun1989backpropagation}, and attention-mechanisms \cite{lim2021temporal}, have shown promising results as they extract complex features and adapt to various time series patterns. These architectures are often designed to learn auto-correlations and cross-correlations from data \cite{zhou2021informer,wu2021autoformer,nie2022time}. 
These kinds of models have successfully integrated exogenous features (e.g. past covariates, known future information, and static covariates) and achieved remarkable success in real-world problems, including traffic forecasting \cite{zhang2022temporal,zhou2023temporal}, retail demand prediction \cite{eisenach2020mqtransformer,salinas2020deepar}, power generation prediction \cite{zhang2024asset}, and energy consumption modeling~\cite{zheng2023interpretable}.

While these models can effectively incorporate numerical or categorical exogenous features, other valuable features like product descriptions or customer reviews exist only as unstructured text. Because this information exists only as unstructured text which is difficult to featurize, these sources of information have been largely neglected or featurized in simple ways \cite{nie2022time, wu2023timesnettemporal2dvariationmodeling, ansari2024chronos, wen2017multi}. 

Nonetheless, descriptive and contextual information about the time series may meaningfully enhance forecast quality. In the retail setting, unstructured text describing the use and design of a product contains valuable information about whether this product will see surges in demand for upcoming seasonal events (holidays, back-to-school, etc.). This information is often difficult or impossible to learn from the time series themselves, for several reasons. Many products are new, with perhaps only one (or \emph{fewer}) years of sales history. Sales history at the product level is also noisy, subject to spurious spikes and stockouts which distort the historical sales. These factors mean that the available historical sales are often insufficient to predict upcoming seasonal patterns reliably. 

\begin{figure}[t] 
    \centering
    \includegraphics[width = 0.75 \textwidth]{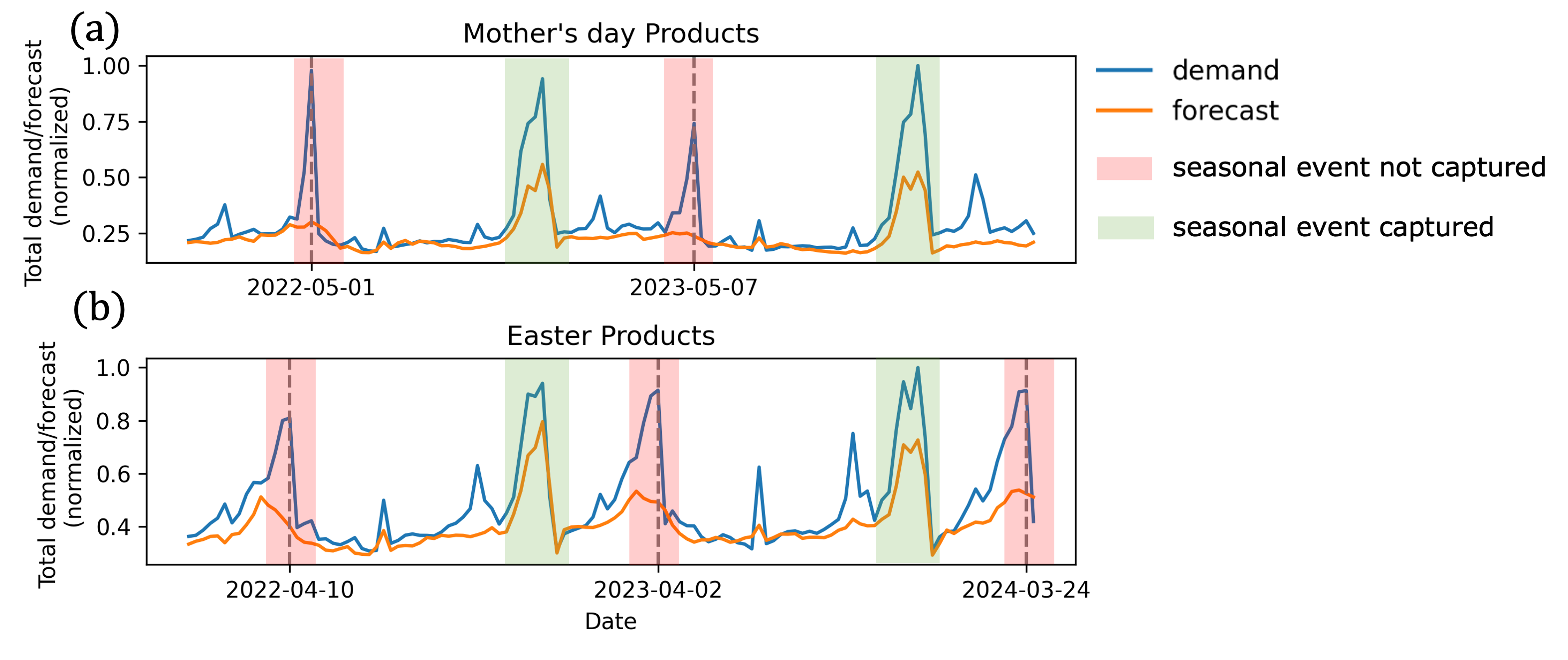}
    \caption{Aggregated demand and forecast for groups of products: (i) Mother's Day products; and (ii) Easter products. The vertical dashed lines mark the week prior to the holiday in question. In green, we show time segments where the production model anticipates event-driven demand surges --- specifically large shopping events like Christmas. In red, we show time segments where the production model fails to anticipate event-driven demand surges.
}
    \label{fig:intro-example}
\end{figure}

In Figure \ref{fig:intro-example}, we illustrate this problem, using forecasts from an MQ-Transformer model \cite{eisenach2020mqtransformer} trained on a retail dataset. 
We focus on products which are relevant to two key holiday related seasonal events: Mother's Day and Easter. 
For both groups of products, we see that the existing forecasting model appropriately anticipates surges in demand during the holiday season (between Black Friday and Christmas). The model responds reasonably to the holiday season (green bands), where we see elevated sales across many products, high customer traffic and numerous price discounts.
By contrast, the model fails to anticipate surges in demand during the Mother's Day and Easter holiday periods themselves (red bands).  Notably, the demand surges during Mother's Day and Easter are localized to small groups of products, aggregate customer traffic does not spike, and products are infrequently discounted. The failure to anticipate these localized surges in demand increases the risk of stockouts for key products during these events. Today, these defects must be addressed through human intervention, relying on human analysts to use their judgment and knowledge to identify products relevant to an upcoming event and adjust the forecasts accordingly. 

Recent advancements in Large Language Models (LLMs) offer a promising avenue to address these challenges, enhancing predictive accuracy by combining rich textual information with traditional covariates. Gruver et al. \cite{gruver2023llmtime} first showed that even LLMs with text-based pre-training can perform impressive zero-shot time series forecasting. Moreover, pre-trained LLMs have the capability to perform domain-specific predictive tasks by directly querying them with domain-specific instructions and knowledge \cite{jiang2024empowering,yu2023temporal,wang2023would,liu2023large}. 
Xue et al. \cite{xue2023promptcast} extended this approach by generalizing prediction tasks to time series data, incorporating context and semantic information from historical data. LLM4TS\cite{chang2023llm4ts} utilizes a two-stage fine-tuning process to improve the model's ability to handle time series data, even with limited data availability. Similarly, TEMPO \cite{cao2023tempo} applies a Generative Pre-trained Transformer (GPT) to time series forecasting, using a prompt-based approach that tailors the model to complex temporal patterns and non-stationary data. Most existing research in time series forecasting has focused on using time series data as input to LLMs, exploring methods like tokenization of time series data \cite{nie2022time,ansari2024chronos}. 

Despite these advancements, it remains an open question how to develop LLMs to integrate descriptive information about the time series themselves --- for example, the description of the product corresponding to the sales in question. Further, these methods are often stand-alone forecasting models. This can be powerful, but in other real-world use cases we may already have a ``good enough'' forecasting solution in place. In those cases, we do not want to completely replace the existing solution --- instead we only want to modify the forecasts to fix areas where the existing models has clear deficiencies.

Our work addresses both of these areas. Here we introduce a procedure, the \emph{LLMForecaster}, to incorporate unstructured textual information and recommend forecast adjustments to improve the accuracy of an existing forecasting pipeline. Our procedure utilizes fine-tuned LLMs that incorporate both historical forecasts as well as unstructured information. This allows us to systematically improve forecast quality in cases where the existing model fails to anticipate holiday related demand surges. We show that our approach enhances the accuracy of demand forecasts in retail environments, empowering businesses to better manage seasonal fluctuations and optimize their operations.

\vspace{-2mm}
\section{Methodology}
\vspace{-3mm}

\begin{figure}[h!]
    \centering
    \includegraphics[width = 0.75\textwidth]{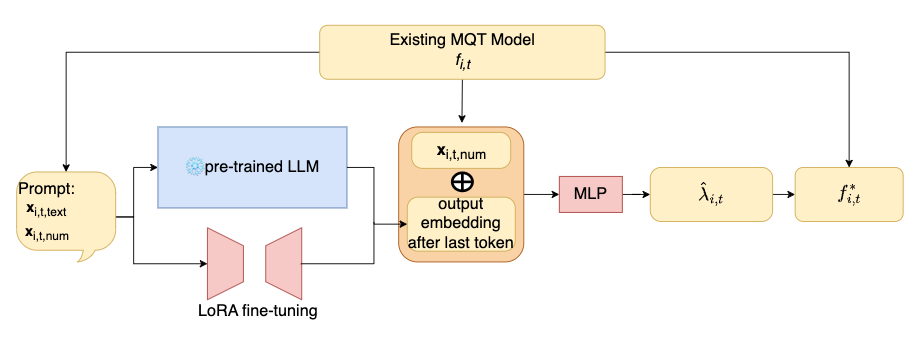}
    \caption{LLMForecaster incorporates text and numeric information through an LLM to rescale the raw prediction $f_{i,t}$, producing a better forecast $f^*_{i,t}$.
    }
    \label{fig:diagram}
\end{figure}

We introduce the LLMForecaster, designed to automate forecast adjustments to correct biases in existing demand forecasting models. 
Here, the existing model is an MQ-Transformer model trained on a large dataset of retail sales. 
This existing model generates initial predictions, denoted as $f_{i, t}$ for product $i$ and target date $t$. 
We then train an additional model, incorporating unstructured text information such as product titles and descriptions ($\mathbf{x}_{i, t, \text{text}}$) and numeric features such as the price or forecast ($\mathbf{x}_{i, t, \text{num}}$) via prompts to an LLM. The model architecture is shown in Figure \ref{fig:diagram}.

This additional model is trained to predict the scaling factor $\lambda_{i, t} = \log (\frac{y_{i, t}}{f_{i, t}})$, where $y_{i, t}$ represents the actual observed demand. We choose to predict a scaling factor, as it allows us to make improvements on top of an existing primary model, and helps deal with heavy-tailed response data. \cite{tripuraneni2023metaanalysisrandomizedexperimentsapplications}. The model minimizes the absolute error between the true and predicted scaling factors:

\begin{equation}
\min_{\hat{\lambda} \in \Lambda} \sum_{i=1}^{n} \big| \lambda_{i, t} - \hat{\lambda}(\mathbf{x}_{i, t, \text{text}}, \mathbf{x}_{i, t, \text{num}}) \big|
\end{equation}
where $n$ is the number of samples, and $\Lambda$ is the model space.

We then use this scaling factor to rescale the original predictions $f_{i, t}$ into the adjusted forecast $f^*_{i, t}$, using the following transformation.

\begin{equation}
f^*_{i, t} = e^{\hat{\lambda}(\mathbf{x}_{i, t, \text{text}}, \mathbf{x}_{i, t, \text{num}}) } f_{i, t}
\end{equation}

The model depends on an input prompt, which contains product information, forecast values, and other contextual information formatting using a predefined template. An example of such a prompt is shown in Appendix \ref{sec:app:prompt}. The fine-tuned LLM generates an embedding vector, which is then adapted to the specific forecasting task using Low-Rank Adaptation (LoRA) \cite{hu2021lora}. This adapted embedding is concatenated with numerical features and fed into a Multi-Layer Perceptron (MLP) head, which outputs the scaling factor $\hat{\lambda}_{i, t}$. By fine-tuning the LLMForecaster on historical forecasts and demand, the model learns to identify patterns in forecast errors and provide accurate adjustments to the primary model's predictions, especially for products with significant holiday demand surges.

\vspace{-2mm} 
\section{Experiment Results}
\vspace{-3mm}

We apply the LLMForecaster to refine predictions made by the existing global model, \texttt{MQ-Transformer} (\texttt{MQT}), at a lead time of 12 weeks. 
This lead time is selected to provide sufficient time to procure inventory in advance of holiday surges. 
The various features are processed into a prompt and fed into the pre-trained \texttt{MPT7b-Instruct} model\cite{MosaicML2023Introducing}, which is further fine-tuned for the forecasting task. 
Performance is evaluated using the weighted $p_{50}$ quantile loss (wQL), defined as: $\text{wQL} = \sum_{i, t} 0.5 |f_{i, t} - y_{i, t}| /  \sum_{i, t} y_{i, t}$ where $i$ is index of product and $t$ is the forecast target date. 

In this experiment, we train a single model capable of calibrating demand predictions across multiple holidays. We focus on five holidays known for strong seasonality: Halloween, Easter, Mother's Day, Father's Day, and Valentine's Day. The training dataset spans 88 weeks, from August 29th, 2021, to April 30th, 2023, including both holiday-related and non-holiday products. A holiday-related product is one with the holiday name in the item name or product description. The test period covers 48 weeks from May 7th, 2023, to March 3rd, 2024, and is divided into five distinct test sets, one for each target holiday. To ensure the LLM knows when events happen, we use a "Holiday-Encoding Prompt" that provides the LLM with the proximity of the target date to the relevant holiday (\ref{sec:app:exp2}).

\begin{figure}[h!]
    \centering
    \includegraphics[width = 0.75\textwidth]{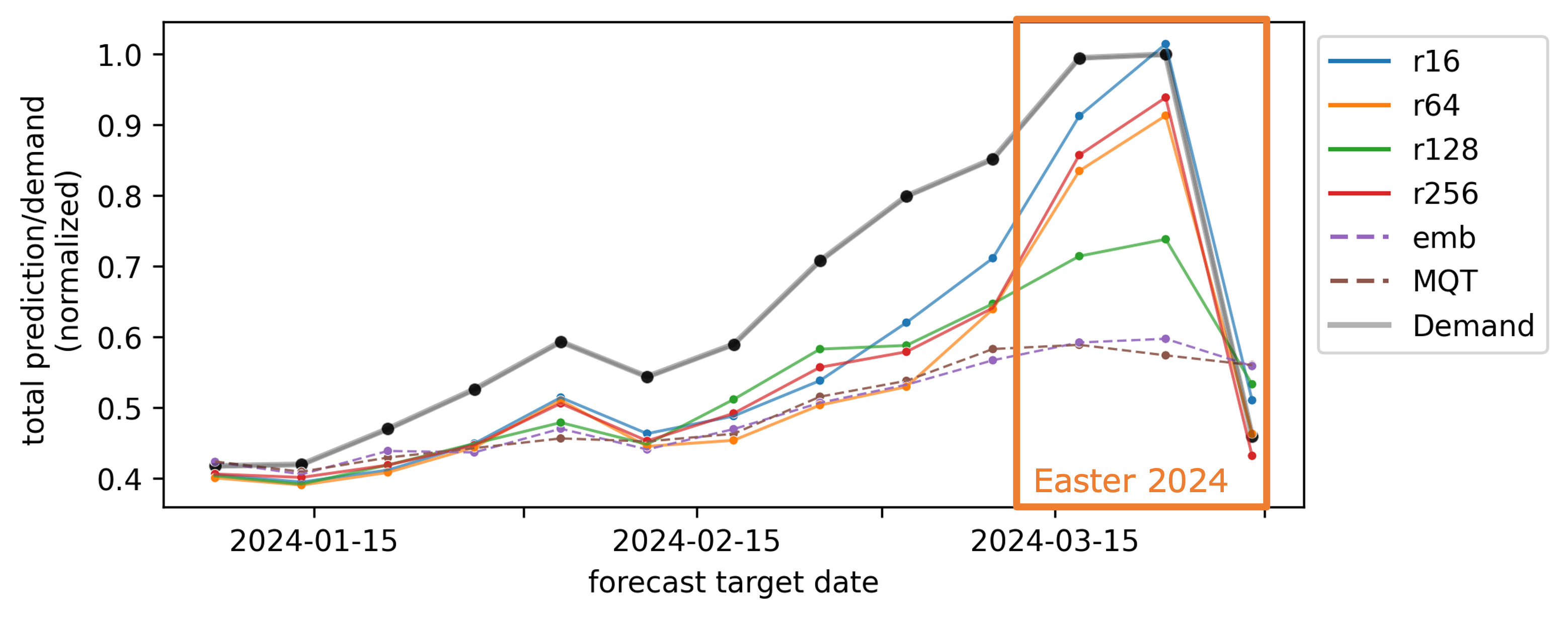}
    \caption{Example of aggregated forecasts on Easter products.}
    \label{fig:Easter example}
\end{figure}

Figure \ref{fig:Easter example} compares the aggregated forecasted demand with actual demand for Easter products. It shows several LLMForecaster models (\texttt{r16}, \texttt{r64}, \texttt{r128}, and \texttt{r256}, representing varying LoRA ranks), the \texttt{emb} baseline (where we do not apply LoRA fine-tuning) and the original \texttt{MQT} baseline. All iterations of the LLMForecaster approach anticipate the Easter demand surge, while our two baselines fail to do so. Table \ref{tab:QLalldates_improvement} presents wQL improvement results for the 48-week test sets demonstrating that the fine-tuned LLMForecaster models (\texttt{r16}, \texttt{r64}, \texttt{r128}, and \texttt{r256}) consistently outperform the baseline \texttt{MQT} and \texttt{emb} models across all five holiday datasets. We also conduct statistical significance testing of the improvement throughout the year - most of these improvements are statistically significant. More detailed empirical results are available in Appendix \ref{sec:accuracy}, and discussion about Valentine's Day are available in Appendix \ref{sec:app-valentines} .

By contrast, the \texttt{emb} model, in which we do not do the LoRA fine-tuning, shows no improvement over the \texttt{MQT} baseline. This underscores the importance of the LLMForecaster's sophisticated fine-tuning approach, leveraging LoRA fine-tuning to effectively learn the holiday-specific demand patterns. 
\begin{table}[ht]
    \centering
    \caption{wQL improvement over the \texttt{MQT} baseline (in basis points) for different testsets \\ \footnotesize{Significance Levels: *** $p<0.001$, ** $p<0.01$, * $p<0.05$} }
    \resizebox{0.7 \textwidth}{!}{
    \begin{tabular}{rllllll}
        Model & Halloween & Easter & Father's Day & Mother's Day & Valentine's Day \\
        \hline
        \texttt{r16}  & $\textbf{105}^{***}$ & $51^{***}$ & $\textbf{93}^{***}$ & $68^{***}$ & $\textbf{58}^{***}$ \\
        \texttt{r64}  & $103^{***}$ & $\textbf{60}^{***}$ & $88^{***}$ & $\textbf{70}^{***}$ & $33$ \\
        \texttt{r128} &  $77^{***}$ & $32^{**}$          & $39^{***}$ & $10^{**}$           & $18$ \\
        \texttt{r256} &  $70^{***}$ & $34^{*}$            & $48^{***}$ & $52^{**}$          & $17^{*}$  \\ \hline
        \texttt{emb}  &  $8$        & $-5$                & $11^{**}$  & $-14$             & $-16$ \\
        \hline
    \end{tabular}
    }
    \label{tab:QLalldates_improvement}
\end{table}

\vspace{-2mm}
\section{Conclusion and Future Work}
\vspace{-3mm}

We introduced the LLMForecaster, a procedure which incorporates unstructured product-level information into numerical time series forecasts and implements forecast adjustments where they are likely to add value; and we demonstrated that the LLMForecaster model leads to statistically significant improvements to product-level demand forecast in large scale backtests in a retail setting.

In future work, we plan to experiment with a broader variety of prompting techniques, as well as hyperparameter optimization. We are actively exploring similar techniques to use LLMs as a tool to featurize data as inputs to deep learning models, rather than as a post-processor. We will also explore multimodal inputs like product images to further enhance forecast accuracy.

\newpage
\bibliographystyle{unsrt}  
\bibliography{references}



\appendix

\section{Appendix}

\subsection{Prompting details}
\label{sec:app:prompt}

The following template is used to apply on the provided text features and numerical features. The part that is being inserted based upon data is indicated with blue square brackets, \textcolor{blue}{[]}. As discussed in \ref{sec:app:exp2}, we also include information if the target date is in the near vicinity of a particular holiday.

\begin{formal}
Pretend you are a sales analyst preparing for the Halloween season. You need to adjust the current model's prediction for a specific product's sales week of \textcolor{blue}{[forecast target date]}.

\#\#\# Instruction:

Today's date: \textcolor{blue}{[forecast creation date]}.

Product Title: \textcolor{blue}{[product title]}

List Price: \$\textcolor{blue}{[list price]}. 

Item Created At: \textcolor{blue}{[item creation date]}. 

Product Group Type:  \textcolor{blue}{[product group]}

Description: \textcolor{blue}{[product description]}

Bullet Points: \textcolor{blue}{[bullet points]}

The current prediction for week \textcolor{blue}{[forecast target date]}'s sales is \textcolor{blue}{[p50]} units with a 90th percentile \textcolor{blue}{[p90]} units. Please provide your adjusted prediction for next week's sales volume considering today's date, the season, the holiday, and the product. Explain your reasoning. 

\#\#\# Response Format: - Prediction: [Your Adjusted Prediction] units - Reasoning: [Explanation] 

\#\#\# Response:
\end{formal}
\subsection{Holiday encoding prompt}

\label{sec:app:exp2}
To improve the LLMForecaster's ability to capture holiday-driven demand patterns, we have implemented a "Holiday-Encoding Prompt". This prompt provides the model with contextual information about the temporal relationship between the target forecast date and surrounding holidays. For example, for a forecast targeting the week of June 1, 2024, the prompt would include:
\begin{formal}
\vspace{-0.2cm}
The mother's day at 2024-05-12 is 3 weeks before 2024-06-01.

The father's day at 2024-06-16 is 2 weeks after 2024-06-01.
\end{formal}

\begin{figure}[h!]
    \centering
    \includegraphics[width = \textwidth]{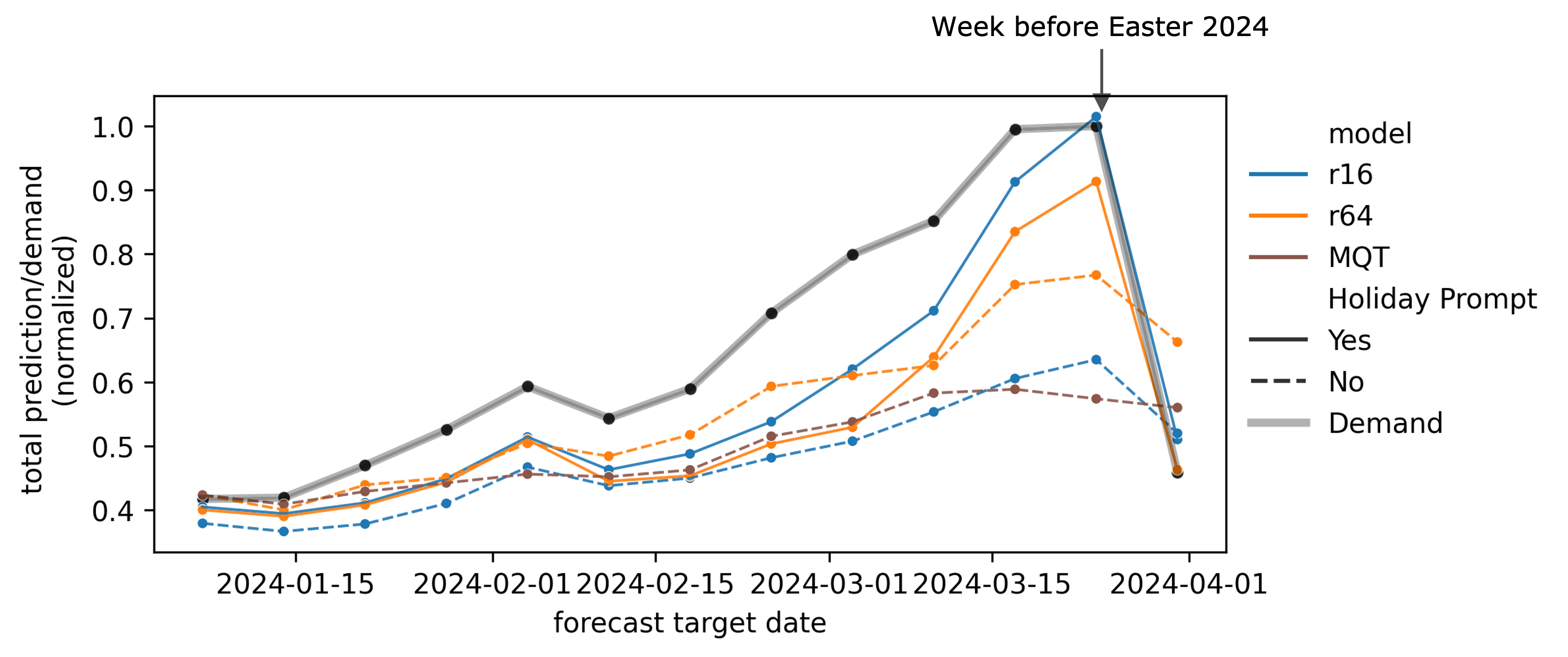}
    \caption{Total demand and prediction for Easter products with and without \textit{Holiday-Encoding Prompt}}
    \label{fig:HolidayPrompt}
\end{figure}

By including these details about the proximity of the target date to relevant holidays, we aim to help the model better identify the appropriate demand patterns. This helps by ensuring the LLM knows precisely when a given event will take place. This is especially important for "moving holidays" like Easter, where the exact date can vary by as much as 35 days from year to year. This Holiday-Encoding Prompt is expected to significantly improve the LLMForecaster's performance in accurately predicting sales for various holiday periods. As shown in Figure \ref{fig:HolidayPrompt}, the proposed LLMForecaster with the Holiday-Encoding Prompt is able to identify the Easter spike at the end of March 2024, while removing the prompt fails to capture the Easter-related demand surge.

\subsection{Accuracy results throughout the year}
\label{sec:accuracy}
Here we show how the LLMForecaster changes forecast accuracy for different groups of products throughout the year. First, we show the results from a t-test measuring the weekly change in quantile loss, over the 48 weeks in the test period. Across the various product sets, we generally see statistically significant improvements over the \texttt{MQT} baseline. The only exception is for Valentine's Day products, which show improvements which are not statistically significant.

\begin{table}[ht]
\centering
\caption{pairwise t-tests of wQL for different testsets and models. \\ \footnotesize{Significance Levels: *** $p<0.001$, ** $p<0.01$, * $p<0.05$ } }
\resizebox{\textwidth}{!}{%
\begin{tabular}{rllllllllll}
 & \multicolumn{2}{c}{Halloween} & \multicolumn{2}{c}{Easter} & \multicolumn{2}{c}{Father's day} & \multicolumn{2}{c}{Mother's day} & \multicolumn{2}{c}{Valentine's day} \\
 \cmidrule(lr){2-3}  \cmidrule(lr){4-5}  \cmidrule(lr){6-7} \cmidrule(lr){8-9} \cmidrule(lr){10-11}
\multicolumn{1}{c}{} & \multicolumn{1}{c}{\texttt{MQT}} & \multicolumn{1}{c}{\texttt{emb}} & \multicolumn{1}{c}{\texttt{MQT}} & \multicolumn{1}{c}{\texttt{emb}} & \multicolumn{1}{c}{\texttt{MQT}} & \multicolumn{1}{c}{\texttt{emb}} & \multicolumn{1}{c}{\texttt{MQT}} & \multicolumn{1}{c}{\texttt{emb}} & \multicolumn{1}{c}{\texttt{MQT}} & \multicolumn{1}{c}{\texttt{emb}} \\
\cmidrule(lr){2-2} \cmidrule(lr){3-3} \cmidrule(lr){4-4} \cmidrule(lr){5-5} \cmidrule(lr){6-6} \cmidrule(lr){7-7} \cmidrule(lr){8-8} \cmidrule(lr){9-9} \cmidrule(lr){10-10} \cmidrule(lr){11-11}
\texttt{r16} & $-5.4^{***}$ & $-5.5^{***}$ & $-3.6^{***}$ & $-3.7^{***}$ & $-5.0^{***}$ & $-3.6^{***}$ & $-3.8^{***}$ & $-4.2^{***}$ & $-3.6^{***}$ & $-4.4^{***}$ \\
\texttt{r64} & $-5.3^{***}$ & $-5.7^{***}$ & $-3.8^{***}$ & $-4.1^{***}$ & $-4.7^{***}$ & $-3.6^{***}$ & $-3.7^{***}$ & $-4.0^{***}$ & $-1.7$ & $-2.2^*$ \\
\texttt{r128} & $-4.2^{***}$ & $-4.4^{***}$ & $-3.4^{**}$ & $-3.6^{***}$ & $-3.9^{***}$ & $-2.3^{*}$ & $-2.4^{**}$ & $-2.7^{**}$ & $-1.9$ & $-2.7^{**}$ \\
\texttt{r256} & $-4.2^{***}$ & $-4.3^{***}$ & $-2.3^{*}$ & $-2.5^{*}$ & $-3.5^{***}$ & $-2.2^{*}$ & $-3.5^{**}$ & $-3.9^{***}$ & $-2.4^*$ & $-3.1^{**}$ \\
\texttt{emb} & $-1.2$ & \multicolumn{1}{c}{-} & \multicolumn{1}{c}{$0.6$} & \multicolumn{1}{c}{-}& $-3.1^{**}$ &\multicolumn{1}{c}{-} & \multicolumn{1}{c}{$0.2$} &\multicolumn{1}{c}{-}& \multicolumn{1}{c}{$2.0$} & \multicolumn{1}{c}{-}\\ \hline
\end{tabular}%
}
\label{tab:ttest}
\end{table}

Next, we show the change in forecast accuracy from our LLMForecaster model versus the \texttt{MQT} baseline. Positive numbers indicate weeks in which the LLMForecaster model improved accuracy over the existing baseline. For each group examined, the LLMForecaster generally improves accuracy throughout the year.

\begin{figure}[h!]
    \centering
    \includegraphics[width = 0.6\textwidth]{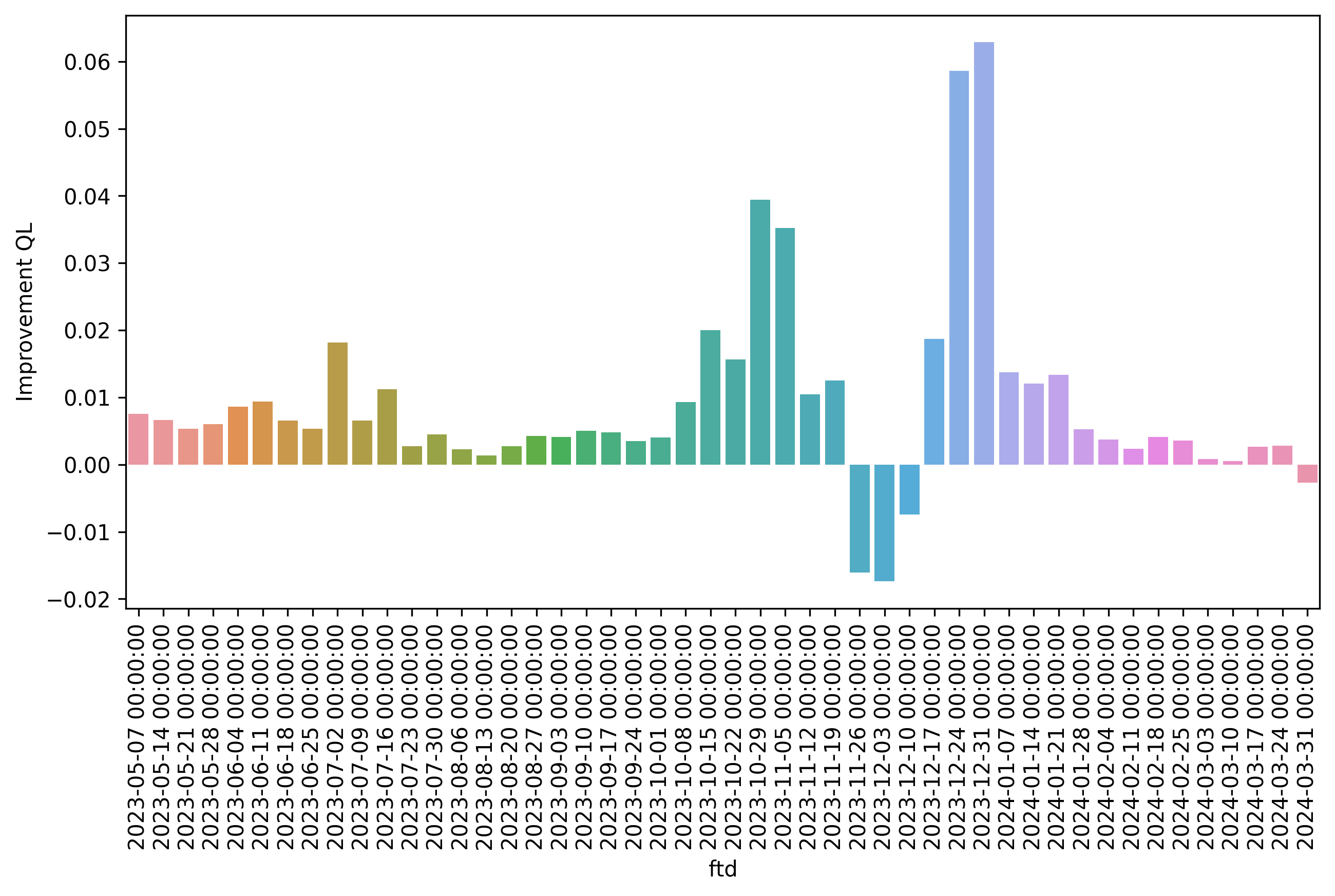}
    \caption{Forecast accuracy change for Halloween products}
    \label{fig:halloween-accuracy}
    \vspace{-4.5mm}
\end{figure}

\begin{figure}[h!]
    \centering
    \includegraphics[width = 0.6\textwidth]{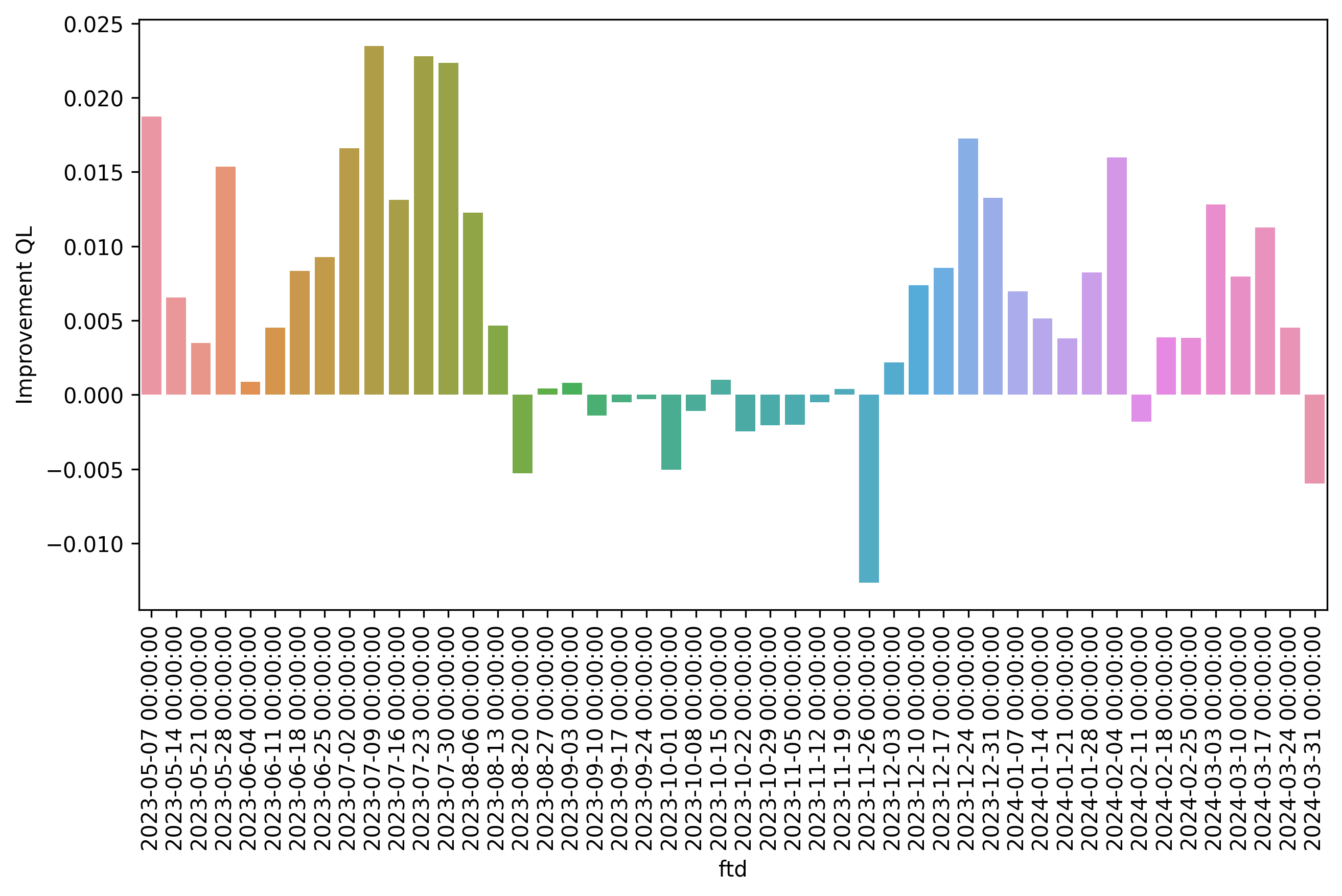}
    \caption{Forecast accuracy change for Mother's Day products}
    \label{fig:mothersday-accuracy}
    \vspace{-4.5mm}
\end{figure}

\begin{figure}[h!]
    \centering
    \includegraphics[width = 0.6\textwidth]{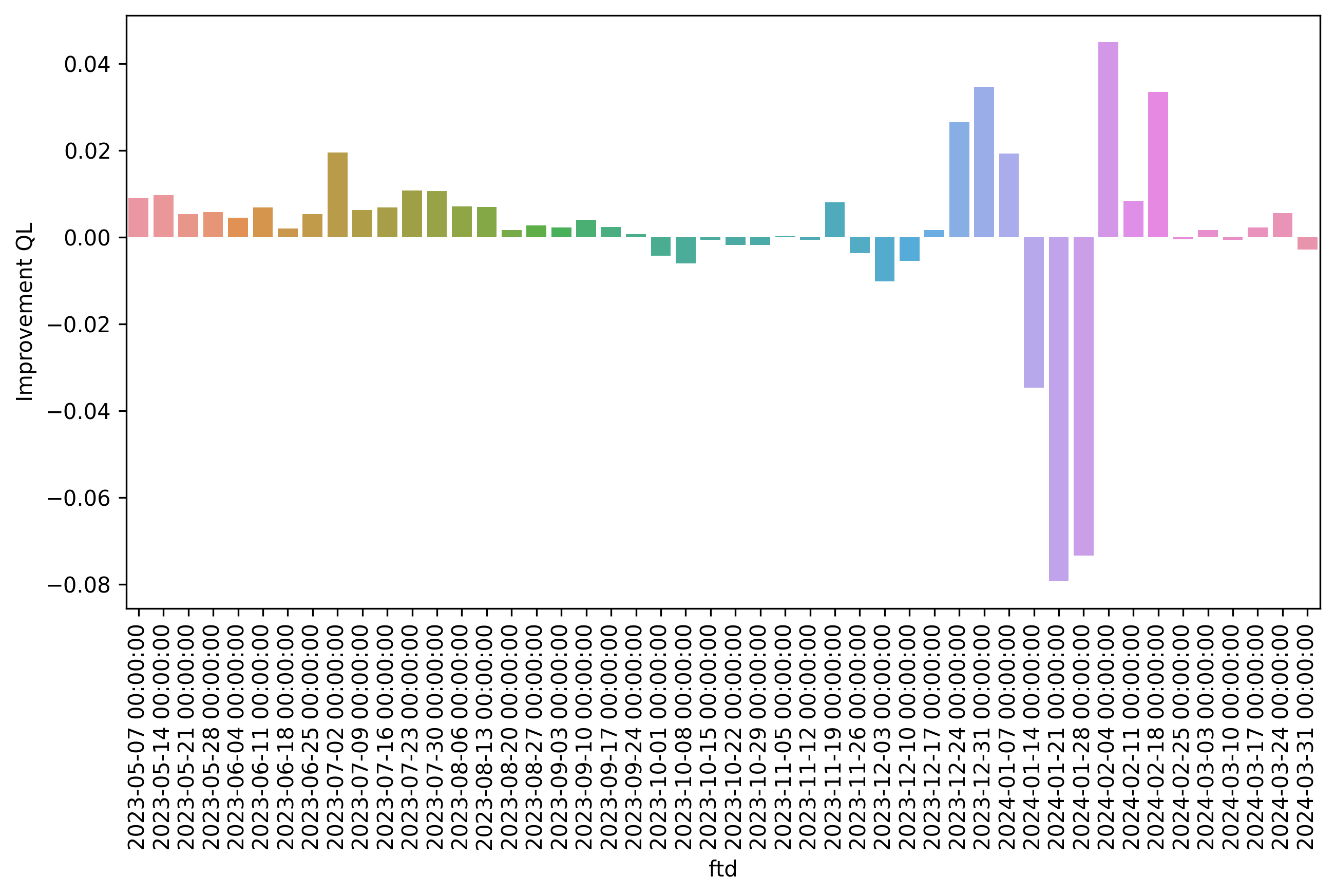}
    \caption{Forecast accuracy change for Valentine's Day products}
    \label{fig:valentines-accuracy}
    \vspace{-4.5mm}
\end{figure}

\begin{figure}[h!]
    \centering
    \includegraphics[width = 0.6\textwidth]{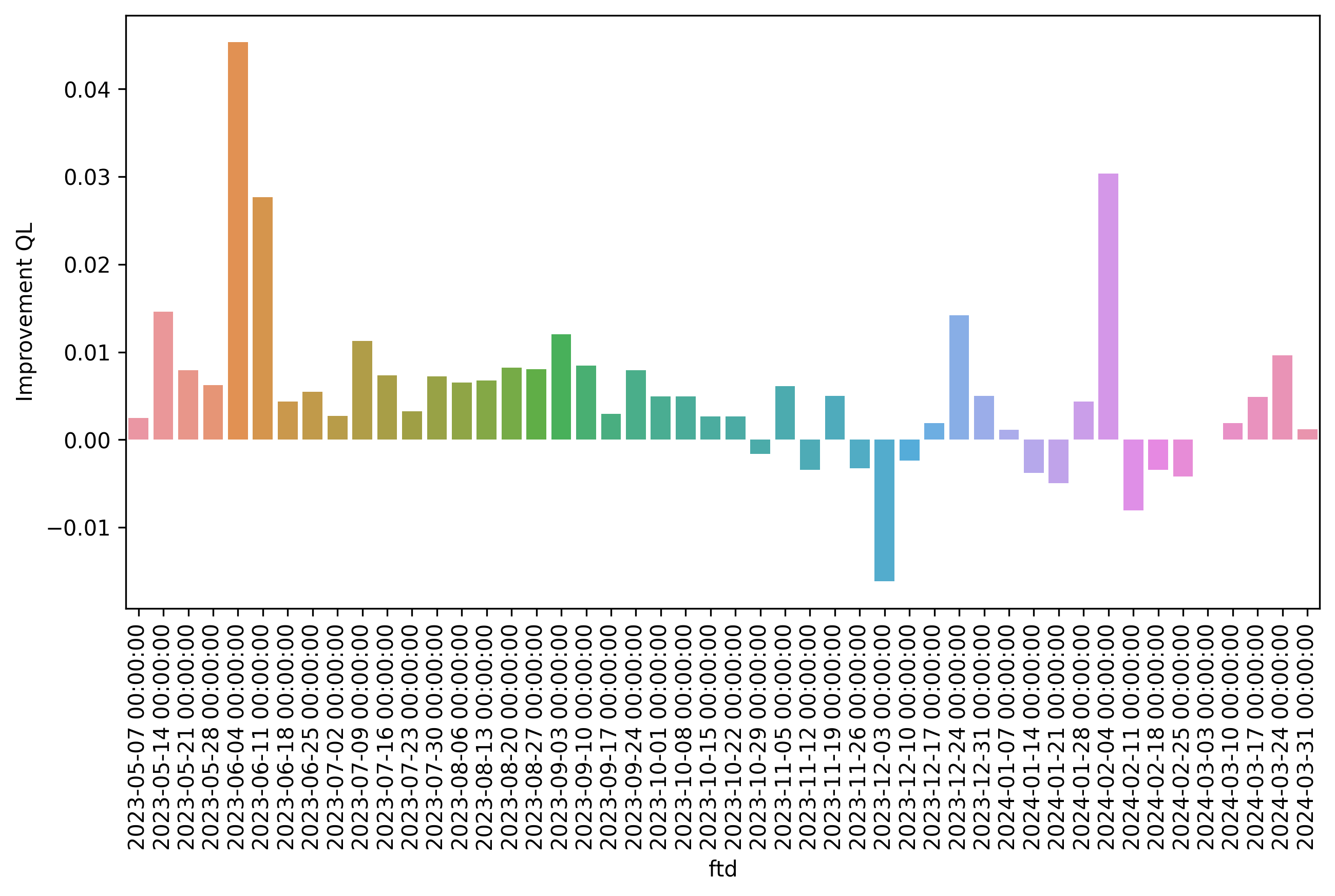}
    \caption{Forecast accuracy change for Father's Day products}
    \label{fig:fathersday-accuracy}
    \vspace{-4.5mm}
\end{figure}

\begin{figure}[h!]
    \centering
    \includegraphics[width = 0.6\textwidth]{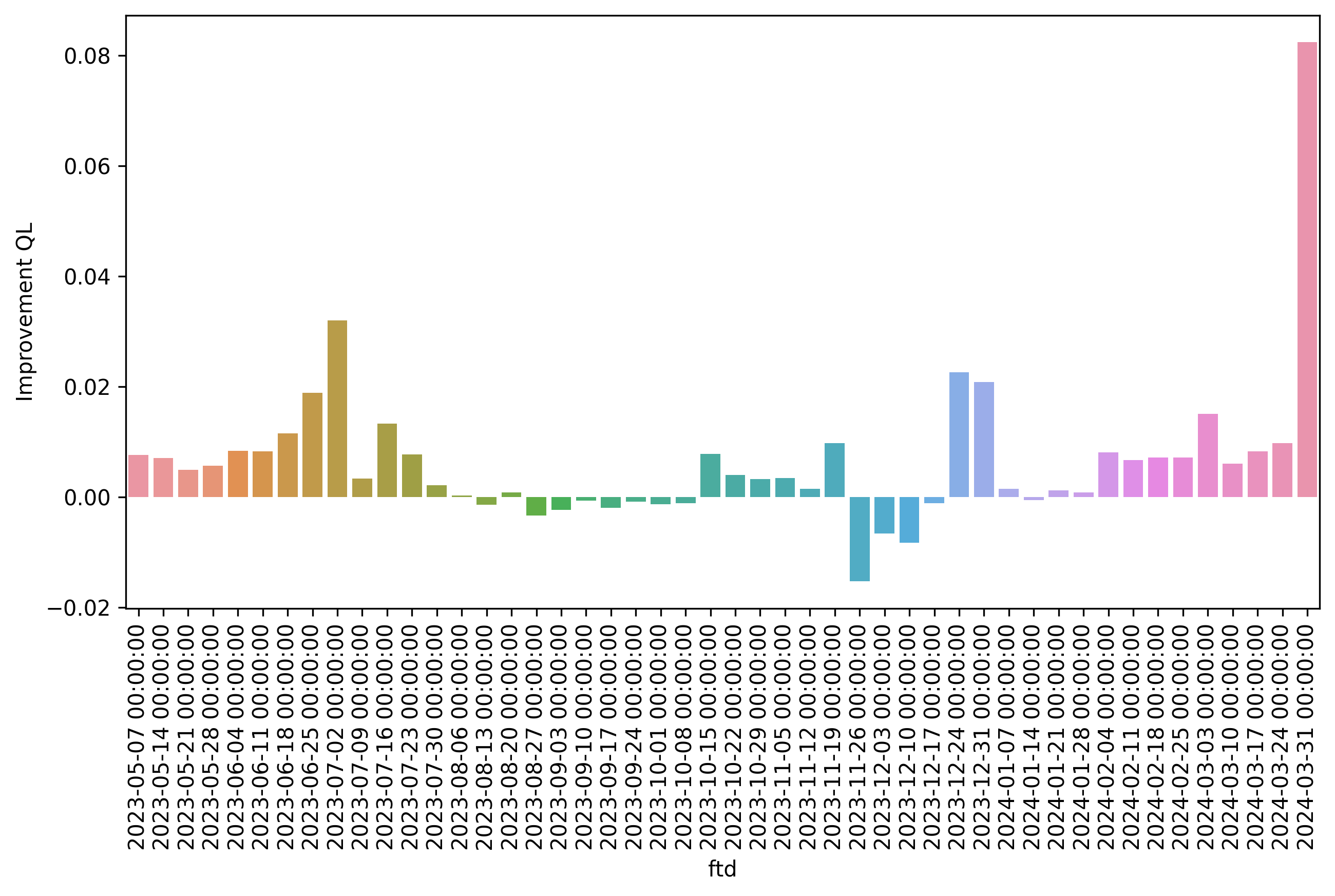}
    \caption{Forecast accuracy change for Easter products}
    \label{fig:easter-accuracy}
    \vspace{-4.5mm}
\end{figure}

\subsection{Valentine's Day}
\label{sec:app-valentines}
\begin{figure}[h!]
    \centering
    \includegraphics[width = \textwidth]{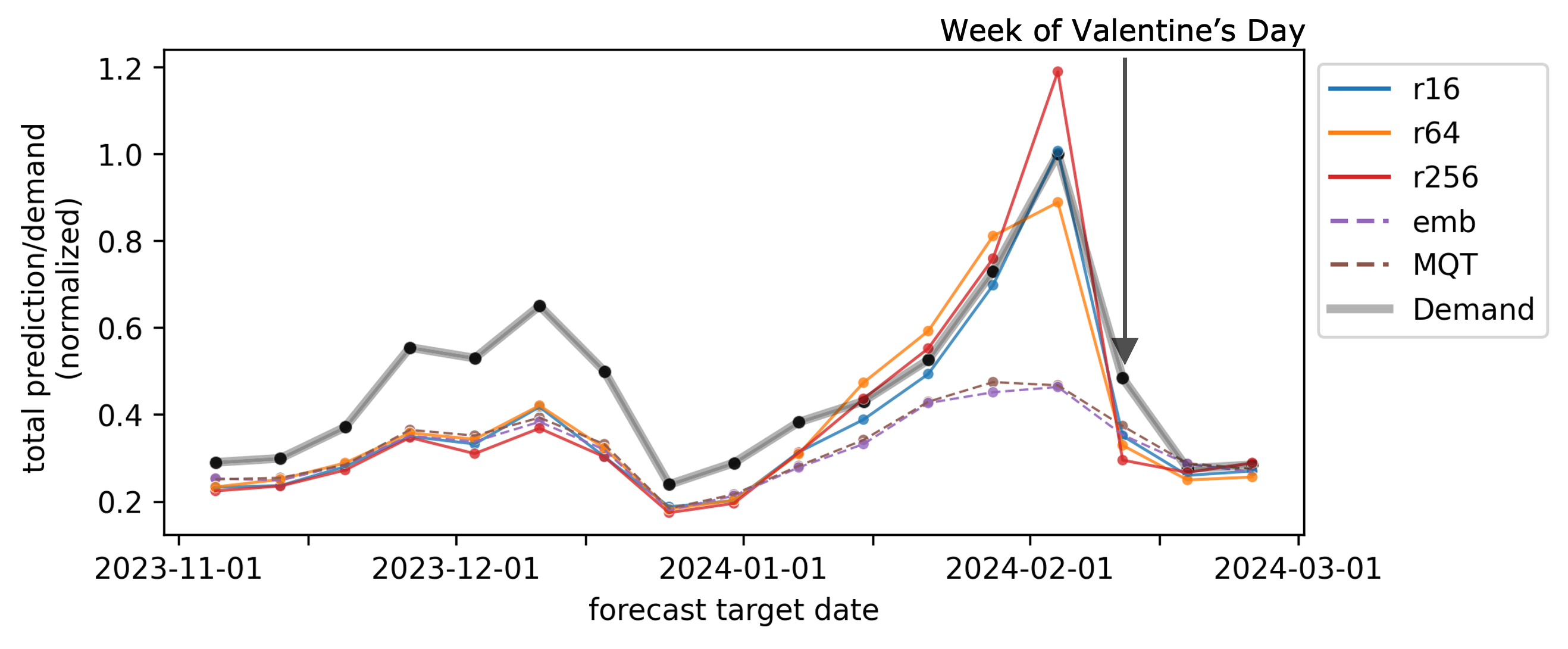}
    \caption{Total demand prediction for Valentine's Day ASINs}
    \label{fig:Valentine's Day}
\end{figure}

The results shown in Tables \ref{tab:QLalldates_improvement} and \ref{tab:ttest} indicate a relatively small and sometimes not statistically significant improvement for the Valentine's Day products compared to the other holiday categories. This can be attributed to the unique nature of the Valentine's Day holiday and its shifting position within the calendar week. For other holidays like Halloween, Easter, Mother's Day, and Father's Day, the dates are fixed on either Saturdays or Sundays, so the overall demand distribution during the holiday period remains relatively consistent. By contrast, Valentine's Day is fixed on February 14th, which can fall on different days of the week. That means that last-minute shopping, for example, may take place in the week of Valentine's Day or the prior week. In this experiment, the training dataset contained Valentine's Day falling on Monday or Tuesday - with minimal time to shop during the week of the holiday, last-minute shopping took place primarily in the prior week. In the test set, Valentine's Day occurring on a Wednesday, so consumers had more time to shop for the holiday during the week itself. When plotting the total prediction and demand on Figure \ref{fig:Valentine's Day}, this effect is clearly observed. While the LLMForecaster models are able to capture the Valentine's Day demand spike compared to the baseline models, they tend to over-predict the weeks before Valentine's Day and significantly under-predict the demand during the actual Valentine's Day week in 2024. This issue could potentially be addressed by training the model with data spanning multiple years, or by incorporating daily demand patterns into the training process. This would help the LLMForecaster better account for the shifting position of Valentine's Day within the calendar week and improve its ability to accurately predict the associated demand patterns.

\end{document}